\documentclass[letterpaper]{article} 
\usepackage{aaai23}  
\usepackage{times}  
\usepackage{helvet}  
\usepackage{courier}  
\usepackage[hyphens]{url}  
\usepackage{graphicx} 
\usepackage{natbib}  
\usepackage{caption} 
\usepackage{algorithm}
\usepackage{algorithmic}

\usepackage{newfloat}
\usepackage{listings}
\DeclareCaptionStyle{ruled}{labelfont=normalfont,labelsep=colon,strut=off} 
\floatstyle{ruled}
\newfloat{listing}{tb}{lst}{}
\floatname{listing}{Listing}
\title{Delving Deep into Pixel Alignment Feature for Accurate Multi-view Human Mesh Recovery}
\author{
    Kai Jia,
    Hongwen Zhang\textsuperscript{$\ast$},
    Liang An,
    Yebin Liu\thanks{Corresponding Authors}
}
\affiliations{

    Department of Automation, Tsinghua University,
Beijing, China \\
    kajia@umich.edu, zhanghongwen@mail.tsinghua.edu.cn, al17@mails.tsinghua.edu.cn, liuyebin@mail.tsinghua.edu.cn
}

\usepackage{bibentry}

\begin{document}

\maketitle

\begin{abstract}
Regression-based methods have shown high efficiency and effectiveness for multi-view human mesh recovery. The key components of a typical regressor lie in the feature extraction of input views and the fusion of multi-view features. In this paper, we present Pixel-aligned Feedback Fusion (PaFF) for accurate yet efficient human mesh recovery from multi-view images. PaFF is an iterative regression framework that performs feature extraction and fusion alternately. At each iteration, PaFF extracts pixel-aligned feedback features from each input view according to the reprojection of the current estimation and fuses them together with respect to each vertex of the downsampled mesh. In this way, our regressor can not only perceive the misalignment status of each view from the feedback features but also correct the mesh parameters more effectively based on the feature fusion on mesh vertices. Additionally, our regressor disentangles the global orientation and translation of the body mesh from the estimation of mesh parameters such that the camera parameters of input views can be better utilized in the regression process. The efficacy of our method is validated in the Human3.6M dataset via comprehensive ablation experiments, where PaFF achieves 33.02 MPJPE and brings significant improvements over the previous best solutions by more than 29\%.
The project page with code and video results can be found at \url{https://kairobo.github.io/PaFF/}.
\end{abstract}

\section{Introduction}

Accurate and efficient recovery of the body mesh underlining a target human is undoubtedly helpful for sub-stream tasks such as behavior understanding~\cite{petrovich2022temos} and human digitalization~\cite{zheng2021pamir}, etc.
With the employment of neural networks, regression-based methods~\cite{kanazawa2018end, kolotouros2019convolutional, guler2019holopose, lin2021mesh} have shown promising results towards this goal. However, regression-based methods typically suffer from coarse alignment between the estimation and the person images. By analogizing the optimization methods~\cite{bogo2016keep, zhang20204d, li20213d}, recent state-of-the-art approaches to monocular human mesh recovery~\cite{song2020human,zanfir2021neural,zhang2021pymaf} make attempts to predict a neural descent to estimate parameters iteratively from feedback signals such as keypoint and part alignment errors~\cite{song2020human,zanfir2021neural}, or re-project the estimation to the original feature space to get pixel-level feedback features~\cite{zhang2021pymaf}. These two trends of methods all utilize feedback features to update the estimation iteratively, which achieves better alignment quality than the previous regression-based methods.
However, the accuracy of these monocular approaches remains far from satisfactory due to the underdetermined observations from a single image.

When deploying regression-based methods on multi-view setups, it is crucial to fuse multi-view information so that complementary observations can be considered in the regression process. Existing multi-view regression-based methods have proposed several fusion strategies, including view-by-view~\cite{liang2019shape,yao2019recurrent}, volumetric~\cite{iskakov2019learnable,shin2020multi}, graph or transformer~\cite{wu2021graph,zhang2021direct,yagubbayli2021legoformer,shuai2021adaptively, he2020epipolar}. View-by-view fusion methods~\cite{liang2019shape,yao2019recurrent} perform estimation view by view and pass the estimation to the next view or stage, which gives improvement from their initial estimations. However, these methods do not consider all of the camera parameters in each stage, which can lead to a large multi-view misalignment. Volumetric fusion methods~\cite{iskakov2019learnable,shin2020multi} utilize back-projection operation to construct a feature volume and then use 3D convolution to fuse the spatial features. Nevertheless, these methods would introduce quantization errors in the discretization of the volume space. Moreover, noisy camera parameters and self-occlusion might lead to the situation that a target voxel in the 3D space corresponds to different body positions in 2D images, making the fusion ambiguous and less effective. Graph or transformer fusion methods~\cite{wu2021graph,zhang2021direct} capture correlation between features from different views to search or infer the best fused feature for the final estimation. But these features are too sparse for the estimation of body mesh such as SMPL~\cite{loper2015smpl}. In this work, we propose to fuse features on mesh vertices and show that such a vertex-wise fusion strategy is more suitable for multi-view human mesh recovery.

\begin{figure*}
\centering
\includegraphics[height=0.85\columnwidth,width=2.1\columnwidth]{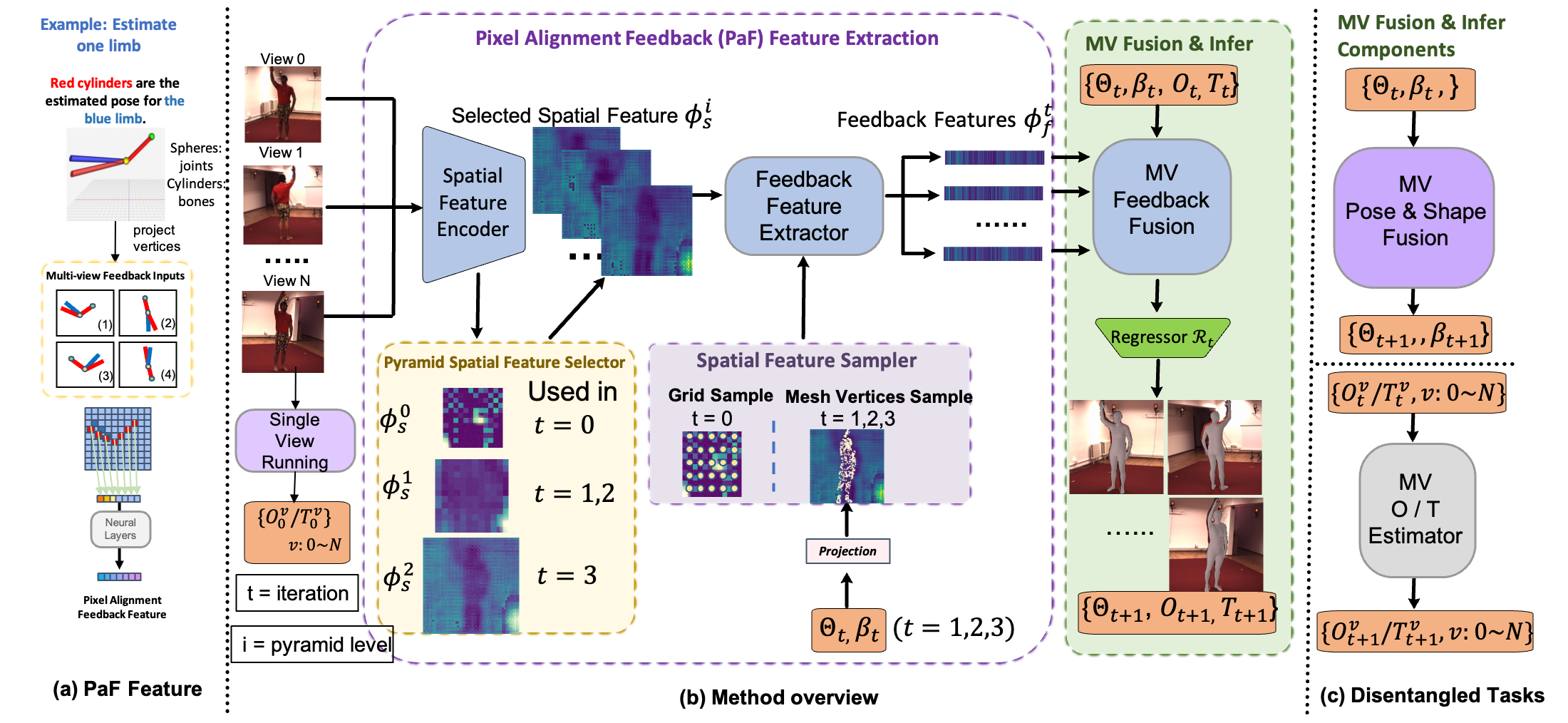}
\caption{Overview of our proposed Pixel-aligned Feedback Fusion (PaFF) pipeline: (a) Pixel Alignment Feedback (PaF) Feature Extraction. (b) PaFF iteratively refines human body parameters' estimation with the guidance of the PaF feature. 
(c) The task of multi-view feedback fusion is disentangled into three tasks - Multi-view Pose \& Shape Fusion, Multi-view Orientation Estimation, and Multi-view Translation Estimation to incorporate camera parameters in the end-to-end model. }
\label{fig:overview}
\end{figure*}

Based on the above motivations, we propose Pixel-aligned Feedback Fusion (PaFF) for multi-view human mesh recovery. PaFF uses the pixel-aligned feedback features in the feature extraction phase and regresses the mesh parameters using the features fused on mesh vertices. As illustrated in Fig.~\ref{fig:overview}(a), the pixel-aligned features are extracted based on the feedback re-projection of the current estimation and reflect the alignment status on each input view.
Fusing the alignment status on mesh vertices could provide more explicit guidance for our regressor to update the mesh parameters.
When extracting the pixel-aligned features, an accurate estimation for orientation and translation is needed since it can align the estimation close to the body region in multi-view images and help us extract more informative feedback features for the pose and shape estimation.
To this end, we disentangle the global orientation and translation from the pose and shape estimation and carefully design the orientation and translation estimators.
Specifically, we incorporate camera parameters in the regression process to figure out the optimized orientation and translation that accord with each input view.
Such a strategy enables our regressor to produce the global orientation and translation in an end-to-end manner and better handle the scale and rotation ambiguity issues.
Different from the previous triangulation-based solution~\cite{iskakov2019learnable}, our method does not depend on keypoint detection results and thus is free from detection mistakes under challenging cases such as occlusions.

We conclude our contributions as follows:
\begin{itemize}
\item We propose Pixel-aligned Feedback Fusion (PaFF) for regression-based multi-view human mesh recovery.
PaFF iteratively extracts pixel-aligned feedback features from each input view and fuses them on mesh vertices.
The feedback features and vertex-wise fusion enables our regressor to update the parameters such that the body mesh is progressively aligned to each input view.
\item We propose to disentangle the global orientation and translation from the estimation of mesh parameters since they are correlated to the camera parameters. 
In this way, the camera parameters can be better utilized in our regressor to overcome the scale and rotation ambiguity issues for a more accurate estimation of the global orientation and translation. 
\item Our method achieves state-of-the-art performances and brings significant improvements over previous methods in benchmark datasets with both calibration and calibration-free settings.
PaFF provides an end-to-end solution for accurate, simple, yet efficient human mesh recovery from multi-view images.
\end{itemize}

\section{Related Work}

The recovery of human body mesh from RGB images has been actively studied in recent years~\cite{bogo2016keep, kanazawa2018end,kocabas2020vibe, caliskan2020multi,zhang2021pymaf, sengupta2021probabilistic, kolotouros2021probabilistic}.
Existing methods can be roughly divided into two paradigms, i.e., optimization-based methods~\cite{liu2011markerless,xu2017flycap,bogo2016keep,li20213d, zhang2021lightweight, huang2017towards,zanfir2021neural, ajanohoun2021multi} and regression-based methods~\cite{pavlakos2017coarse,omran2018neural, kanazawa2018end, varol2018bodynet, kolotouros2019learning,rong2019delving,rong2021frankmocap,guler2019holopose,kolotouros2019convolutional,zhang2020learning,liang2019shape,sun2021monocular,zhang2021pymaf,lin2021mesh,xuan2022cicai}. 
We refer readers to~\cite{tian2022recovering} for a comprehensive survey in this field.

Regression-based methods for multi-view human mesh recovery~\cite{liang2019shape, shin2020multi, sengupta2021probabilistic, zhang2021direct} usually need to go through a feature extraction phase, a multi-view feature fusion phase, and an inference phase. To fuse the information from the multi-view,~\citet{liang2019shape} proposes to estimate a human body stage by stage and view by view without using camera parameters. 
~\citet{shin2020multi} fuses the multi-view features to a 3D feature volume using back-projection and then regresses the body parameters from the flattened volumetric feature, giving the state-of-art accuracy in the multi-view human body reconstruction, yet with quantization errors and possible occluded non-body regions incorporated.

While optimization-based methods~\cite{bogo2016keep, huang2017towards, li20213d, ajanohoun2021multi} can fit the estimation aligned to 2D evidence iteratively, regression-based methods~\cite{kanazawa2018end, kolotouros2019convolutional} usually suffer from bad alignment to the human image region. Recently, Neural Descent methods mimicking optimization processes appeared in the human body estimation task~\cite{carreira2016human, zanfir2021neural, song2020human, zanfir2021neural, corona2022learned}. These methods use the iterative regression with feedback signals such as keypoint re-projection errors and body part alignment errors~\cite{zanfir2021neural} to infer a neural descent of the parameters. Nevertheless, these methods can be sensitive to noisy 2D evidence. Recently, several methods extract the implicit feedback signals from the image features by projecting estimation to the image area~\cite{zhang2021pymaf, zhang2021direct}, which contains richer alignment information.~\citet{zhang2021pymaf} uses the mesh re-projection feedback feature while~\citet{zhang2021direct} projects estimated 3D keypoints into each view and utilizes deformable convolution~\cite{zhu2019deformable} to extract efficient feedback features.  Compared with the previous methods only relying on numerical errors~\cite{song2020human, zanfir2021neural}, these methods can utilize the rich image features to get a more aligned estimation and have more information in tasks such as shape estimation. It also relies less on the on-the-shelf models such as keypoints detection or part segmentation which might introduce additional noises. However, these methods are either not applied in the multi-view setting~\cite{zhang2021pymaf} or only used for the human 3D keypoints estimation task~\cite{zhang2021direct}. Therefore, an efficient solution of the multi-view human reconstruction with the feedback loop is meaningful.

\section{METHODOLOGY}
\subsection{Overview}
Given $N$-view images of single human body and camera parameters $\{K^v_{cam}$, $R^v_{cam}$, $T^v_{cam}, v=0,1,...,N\}$, the multi-view human mesh recovery task requires estimating the body parameters (typically SMPL~\cite{loper2015smpl} pose parameter $\theta$ and shape parameter $\beta$), the global translation $T_g$, and the global orientation $O_g$ simultaneously.
To solve this problem effectively, 
our proposed method iteratively refines all the parameters with pixel alignment feedback (PaF) features, as illustrated in Fig.~\ref{fig:overview}(b).
Specifically, a pre-trained single-view pixel alignment feature extractor runs first to extract an image feature pyramid $\{\Phi^{i}_{s}, \space i=0,1,2\}$ and gives the initialization estimation of body orientation $O_0^v$ and translation $T_0^v$ for each view. During multi-view fusion, PaFF adapts the Spatial Feature Sampler to sample multi-view feedback features $\Phi^{t}_{f}$ on the collected feature pyramid, followed by a multi-view feedback fusion module to aggregate all the feedback features in order to infer parameter updates for each iteration.
Note that our model could work in arbitrary numbers of camera views given the camera parameters, yet $N=4$ is used in the paper. There are 4 iterations of PaFF multi-view regression.

\subsection{Multi-view Pixel Alignment Feedback Feature}
The iterative regression process, which estimates the residual estimation updates iteratively to get the final estimation, has been proven to be beneficial for accurate human mesh recovery~\cite{kanazawa2018end, kolotouros2019convolutional, shin2020multi}.
However, previous methods usually utilize feed-forward iteration, which means all iterations only engage the same global feature vector in each iteration. 
Here comes a critical drawback: decoding pose and shape from a global feature vector is hard to achieve pixel-aligned performance no matter how many iterations are performed. In order to obtain pixel-aligned performance, we seek to add image information to each iteration which serves as the \textit{feedback}.
To realize it, we make crucial improvements for multi-view feature extraction by projecting the estimated human mesh vertices to the image plane and constructing pixel-aligned feedback (PaF for short) features by concatenating sampled features from an intermediate feature map, as shown in Fig.~\ref{fig:overview}(a). As the previous study~\cite{dijk2019neural} shows, the pixel position of a feature can be encoded by the neural network.
We believe the relative position between the sampled pixels and the pixels inside the real human body region can be encoded into the feedback features and inform the misalignment between the estimated and the real human body in each camera view. 

As illustrated in Fig.~\ref{fig:overview}(b), we construct a PaF Feature Extraction Model, in which a Spatial Feature Encoder extracts a coarse-to-fine feature pyramid with three feature maps and a Spatial Feature Sampler samples estimated mesh vertices from the feature maps to get PaF features.
During the initial iteration, where no previously estimated vertices are available, we apply grid sampling on the first feature map $\Phi^{0}_{s}$ to extract the initial point-wise features. There are three additional iterations to align the estimated body with multi-view 2D evidence. During these three iterations, we adopt a Mesh Vertex Sampling method to extract the PaF features from feature maps $\{\Phi^{i}_{s}, \space i=1,1,2\}$ for iteration $t=1,2,3$. Note that $\Phi^{1}_{s}$ is reused in iterations 1 and 2. Gradually, these feedback features would guide the sampled points closer to the true human area with an updated parameter set.

To make the extracted feedback feature more effective and generalized, we pre-train the Pixel-alignment Feedback Extraction Model with the existing 2D datasets by concatenating a monocular inference module. The pre-trained model also takes four iterations to regress the final prediction. $O^v$ and $T^v$ estimations from the last iteration of each view's running would be reserved to initialize the multi-view estimation to make the body projection to align faster. Details of the pre-training are shown in the supplementary material. 

\subsection{Multi-view Feedback Fusion Module}
After the extraction of PaF features from multi-view images, a multi-view fusion module will compile the misalignment information of different views before the final inference. 
As shown in Fig.~\ref{fig:overview}(a), one view's misalignment information of the limb is not adequate to correct false body parts due to depth and shape ambiguity. Therefore, the goal of the feedback feature fusion is to leverage the multi-view misalignment information in one fused feature fully. 
Different from existing multi-view methods~\cite{shin2020multi, zhang2021direct} which fuse a single multi-view feature to infer all the parameters ($\theta, \beta, O_g$ and $T_g$), we disentangle the multi-view fusion and inference into three isolated tasks - i) pose \& shape estimation, ii) global orientation estimation, and iii) global translation estimation. 
The benefit of the disentangling is to separate the inferring into two groups, body global pose (orientation and translation) and body local pose (change of joints and shape) in the iterative feedback regression process,
which encourages the different modules to focus on their own task scale.

\subsubsection{Multi-view Pose and Shape Fusion}

\begin{figure}
\centering
\includegraphics[height=0.454\columnwidth,width=1\columnwidth]{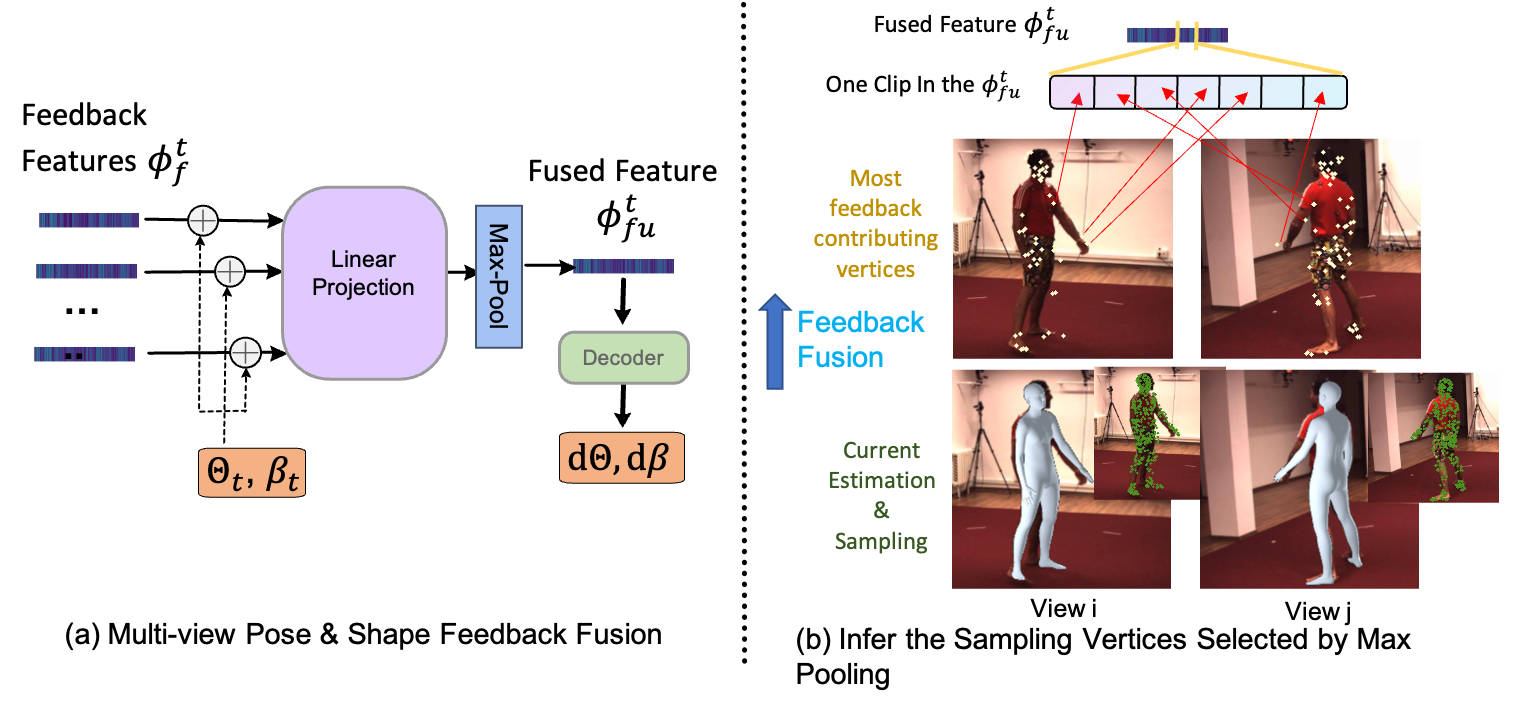}
\caption{Multi-view Pose \& Shape Feedback Fusion Module:
a) The structure of pose \& shape multi-view feedback fusion Module. b) Visualize vertices selected by max-pooling: The estimated meshes from the previous iteration and the sampled vertices are shown at the bottom.
In the top images, we visualize the sampling vertex in each view that contributes most to one feature dimension of the fused feature. By looking at the selected vertices on the left arm and legs, we can find that the vertices which reflect more estimation misalignment are easier to be chosen. }
\label{fig:pose_shape}
\end{figure}
The PaF feature for pose and shape estimation contains information about the misalignment between the true body and the estimated body.
Pose and shape estimations are highly entangled with each other since changes in joint angles or shape parameters can both result in a misalignment, in which bone's skew and shape vertices offset can be hard to distinguish.
The angular misalignment from one view is not adequate to infer the true angular misalignment in 3D space, as shown in Fig.~\ref{fig:overview}(a),
while the true shape misalignment could be ambiguous in one view due to depth ambiguity.
As shown in Fig.~\ref{fig:pose_shape}(a), we first use a linear projection layer to adapt the PaF feature for the pose and shape fusion task.
With a belief that maximum feedback values reflecting more essential misalignment information and non-maximum values might contain noise induced by occlusion, to infer complete misalignment information from multi-view feedback features, we propose to apply max-pooling on the stacked multi-view PaF features. Fig.~\ref{fig:pose_shape}(b) shows the selected  sampling vertices for the fused feedback features, which illustrates large misalignment information can be retained in the process. 
However, the max-pooling operation would inevitably filter some misalignment information contained in the non-maximum signals. The iterative update process can be seen as a solution for completing the misalignment information gradually.

\subsubsection{Global Orientation, Translation Estimator}

\begin{figure}
\centering
\includegraphics[height=0.454\columnwidth,width=1\columnwidth]{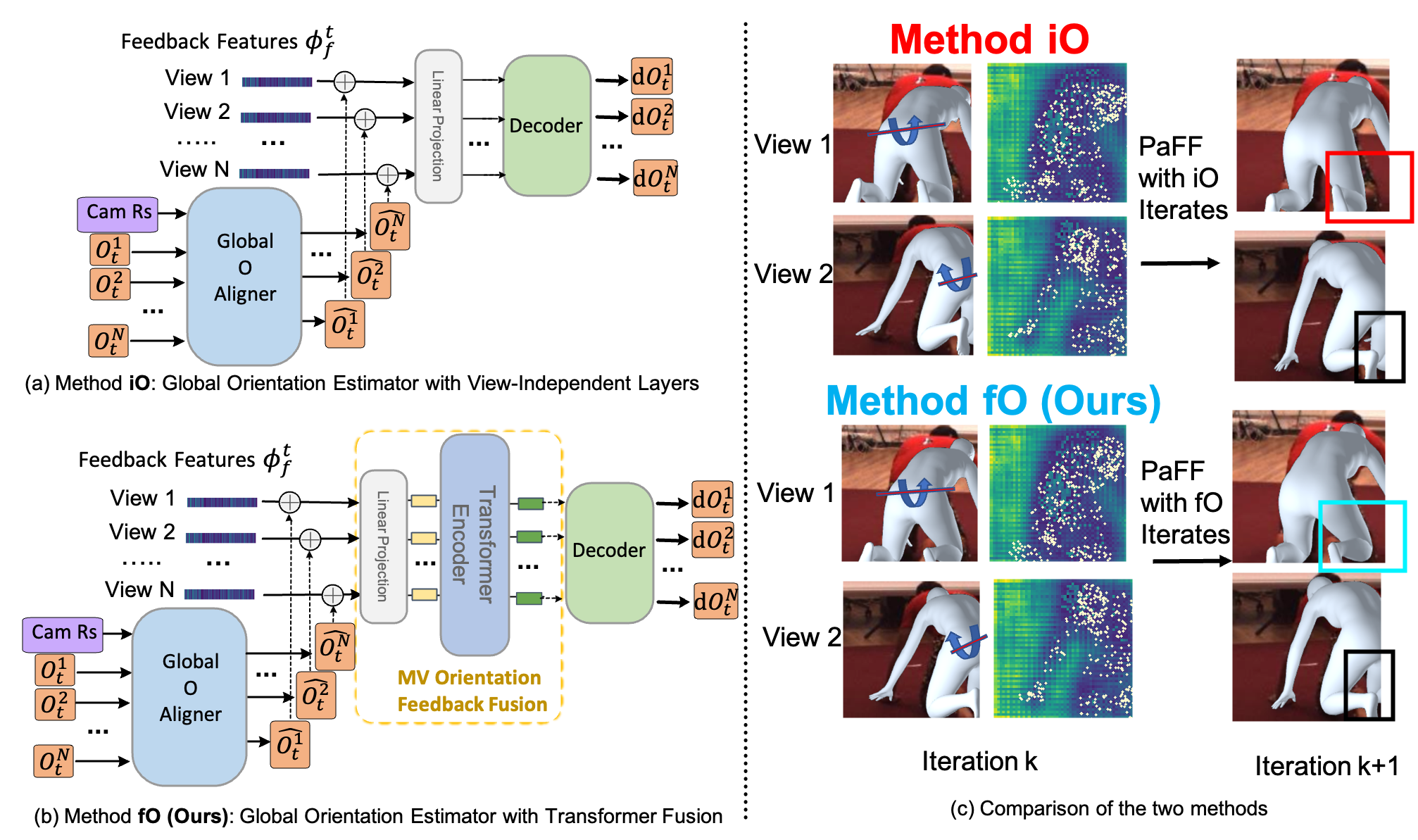}
\caption{Global Orientation Estimator: (a) Method iO and (b) Method fO are two options for Global Orientation Estimator; (c) shows the motivation of method fO that uses transformer fusion to collect multi-view body parts misalignment signals in order to avoid depth ambiguity and occlusion problems.  Estimated models and feature sampling are visualized in the left images, while updated estimations are shown in the right images. Blue arrows indicate the correction rotation needed for each view's estimation. Comparing the refinement effect in one iteration (k=2) of the two methods, method fO performs better in orientation estimation, which also leads to a better estimation of the right leg.}
\label{fig:O_estimator}
\end{figure} 

Different from other multi-view regression-based methods~\cite{liang2019shape, shin2020multi}, our method utilizes a feedback feature - PaF to refine the estimation of orientation and translation iteratively in an end-to-end manner.
The orientation and translation are `optimized' at the same time with pose and shape estimations yet with a limited number of iterations. An accurate estimation for orientation and translation is crucial to align the estimation to the multi-view body regions in order to extract informative feedback signals for local pose and shape updates. Following the monocular estimation methods for human reconstruction~\cite{kanazawa2018end, kolotouros2019convolutional, zhang2021pymaf} to predict approximated camera parameters with the orthogonal projection assumption, we build a camera calibration-free version of our method, which has two independent neural networks for single-view body orientation estimation $O^v$ and relative translation estimation $TO^v$ using a default focal length while the shape and pose are still predicted as a joint estimation from multi-view feedback fusion. 
 When calibrated cameras are given, global orientation $O_g$ and global translation $T_g$ can be estimated to rectify each view's camera prediction ($O^v$ and $TO^v$).
Specially, we choose not to update O and T after grid sampling and use the initial estimations since the grid points do not reflect any misalignment information.
 
The body orientation can be seen as a `root joint' of the human body. One change of orientation can skew all body parts with image evidence, which gives us the intuition of using pixel alignment features to infer the skew of orientation estimation.
As illustrated in Fig. ~\ref{fig:O_estimator}(b), we first align the initial single-view orientation estimations with a Global Orientation Aligner Algorithm using camera rotations. The Global Orientation Aligner Algorithm first filter the most orientation-skewed view and update the orientation estimation for each view as $\widehat{O^v_t}$ with camera rotation parameters (check supplementary for the details).
After the Global Orientation Aligner,
PaF features are extracted with the updated orientation estimations.
Inside the PaF feature, the information reflecting the orientation misalignment mainly comes from the overall estimated body parts' skewing.
Due to depth ambiguity and occlusion (which can be seen in Fig.~\ref{fig:O_estimator}(c)), black and red boxes), one view's body parts' skewing might not be complete for the orientation correction.
So, we design a transformer-based orientation feedback fusion module to capture the co-relationship between multi-view orientation misalignment and complete each view's orientation misalignment information by the attention mechanism.
We construct the fusion module with a multi-head transformer encoder with concatenated multi-view feedback features, and single-view orientation estimation as input queries~\cite{vaswani2017attention}. 
Since the angle of the correction rotation for each view is the same (multi-view features are fusible), while the rotation axis is different (as shown in Fig.~\ref{fig:O_estimator}(c)), we use each view's updated PaF feature to infer a correction rotation for each view independently. 
As shown in Fig.~\ref{fig:O_estimator}(b), multi-view PaF features would go through the transformer encoder to predict orientation update $dO_t^v$ in the 6D representation rotation~\cite{zhou2019continuity}.
We first update $O^v$ for each view in an addictive fashion, i.e., $O^{v}_{t+1} = O^{v}_t + dO_t^v$ and then apply the Global O Aligner to re-align the estimations with camera rotations. A non-transformer version of global O estimator in Fig.~\ref{fig:O_estimator}(a) is compared with our method to illustrate the effect of transformer-based multi-view fusion in Fig.~\ref{fig:O_estimator}(c).

The global translation of the human body $T_g$ can be estimated by triangulation based on keypoint detections~\cite{iskakov2019learnable, tu2020voxelpose} or be ignored by predicting offsets from a template model~\cite{lin2021mesh}. However, keypoints detection can be noisy with occlusion and require an additional off-the-shelf model, and translation estimation is needed for our re-projection process.
As the 2D image feature is hard to reflect perspective camera pose, we are following the monocular methods, which estimate an orthogonal camera of the human body~\cite{kanazawa2018end, kolotouros2019convolutional, zhang2021pymaf} to predict a relative translation and a body scale for each view.
As the relative translation from an orthogonal camera can lead to a scale ambiguity, we solve a global scale by using the camera information. Specifically, we assume the estimated body is aligned after independent orthogonal camera prediction. Then the global location of 3D pelvis and the pelvis 2D keypoint of the estimated body lies in the same camera ray for each view. Since the scale of the human body is the same given by a single shape parameter, we solve an adaptive global scale for the estimated body by estimating a global translation with camera parameters from a linear equation. The adaptive scale is helpful to rescale the body mesh, which will improve the accuracy of the absolute keypoints location estimation.
(check derivation details in supplementary)


\subsection{Training}
There are two stages of training. The first stage is to pre-train the Pixel Alignment Feedback Feature Extraction Model with some monocular human capture datasets. In the second stage, the feature extraction module is fixed since the feature is informative enough after the pre-training process. And we train the multi-view fusion modules without the global translation estimation.
After the second stage, the body alignment for multi-view images 
 is relatively accurate so that the Global Translation Estimator can be applied to solve scale ambiguity. We use 2D/3D keypoints, joint angles, and shape parameters as the supervision signals and add regularization if shape data is not available. (details in supplementary)

\section{EXPERIMENTS}

\begin{table*}
  \centering
  \begin{tabular}{lllllllll}
    \hline
    Methods     & MPJPE     & PA-MPJPE & Multi-view & Cameras & Parametric \\
    \hline
    Deep Triangulation~\cite{iskakov2019learnable} & \textbf{20.8}*  & -  & Yes & Known & Keypoints  \\
    \hline
    SPIN~\cite{kolotouros2019learning} & 62.5  & 41.1  & No & Not Known & Yes   \\
    I2L-MeshNet~\cite{moon2020i2l} & 55.7*  & 41.1*  & No & Not Known & Vertices   \\
    Mesh Graphomer~\cite{lin2021mesh} & 51.2*  & 34.5*  & No & Not Known & Vertices   \\
    PyMAF~\cite{zhang2021pymaf} & 57.7  & 40.5  & No & Not Known & Yes \\
    \hline
    MuVS~\cite{huang2017towards} & 58.2 & 47.1  & Yes & Known & Yes   \\
    Shape Aware~\cite{liang2019shape} & 79.9  & 45.1  & Yes & Not Known & Yes \\
    \citet{shin2020multi} & 46.9 & 32.5   & Yes & Known & Yes   \\
    ProHMR\cite{kolotouros2021probabilistic} & 62.2  & 34.5  & Yes & Not Known & Yes 
    \\
    \citet{yu2022multiview} & -  & 33.0  & Yes & Not Known & Yes 
    \\
    Calib-free PaFF(Ours)  & \textbf{44.8} & \textbf{28.2} & Yes & Not Known & Yes \\
    PaFF(Ours) & \textbf{33.0} & \textbf{26.9} & Yes & Known & Yes \\
    \hline
  \end{tabular}
    \caption{Result on the Human3.6M. Calib-free means calibration-free. * denotes the results of non-parametric representations. } \label{tab:h36m}
\end{table*}

The experiments are designed to answer the following questions. 1). How is the performance of PaFF? Does it have a generalization ability?
2). How do the Global Orientation Estimator and the Global Translation Estimator boost the performance?
3). How does the individual component choice affect the performance, and why?

\noindent{\textbf{Datasets and evaluation metrics}}
We trained our PaF feature extractor on Human3.6M~\cite{ionescu2013human3}, MPI-INF-3DHP~\cite{mehta2017monocular}, COCO~\cite{lin2014microsoft}, MPII~\cite{andriluka20142d}, LSP~\cite{johnson2010clustered}, LSP Extended~\cite{andriluka20142d} using monocular images for the first training stage. Then we freeze the PaF feature extractor's weight and train the rest of the model on Human3.6M and MPI-INF-3DHP with extrinsic and intrinsic camera parameters (For calibration-free PaFF, we do not use them). The train/test split for Human3.6M and MPI-INF-3DHP follows the previous multi-view estimation works~\cite{liang2019shape, shin2020multi}. To evaluate Human3.6M, we remove MPI-INF-3DHP during training since the 3D keypoints ground truth in the dataset is relatively inaccurate.
To evaluate MPI-INF-3DHP, we keep MPI-INF-3DHP during training but would constrain some abnormal losses given by noisy 3D data.  For the evaluation Human3.6M, we use MPJPE, PA-MPJPE, and PVE as the evaluation metrics following \cite{liang2019shape}. For the evaluation of MPI-INF-3DHP, we use the same metrics - MPJPE, PCK and AUC with~\citet{liang2019shape} and~\citet{shin2020multi}. To evaluate the body orientation estimation, we use the angle error between the estimated orientation rotation matrix and the ground truth matrix, which denotes as `O Err'.
To show the generalization ability of PaFF, we additionally train PaFF on MTC Dataset~\cite{xiang2019monocular} by mixing training with Human 3.6M, of which results will be shown in the supplementary.

\noindent{\textbf{Implementation details}}
We built the Spatial Feature Encoder in the PaF feature extractor upon ResNet-50~\cite{he2016deep} and the Feedback Feature Extractor in the PaF feature extractor with 1D convolution layers to downsize the sampled feature. The Multi-view Orientation Fusion Module is constructed by a multi-head transformer encoder with 5 heads and 2 layers. The decoders are all constructed by fully connected layers. We use Adam~\citet{kingma2014adam} optimizer with a fixed learning rate 1e-5. The first stage of training takes 30 epochs with batch\_size of 64 to learn an effective feedback feature extraction. The second stage of training takes 10 epochs. The batch\_sizes for the second stage is 16, and the number of views is 4.
For more details about implementation details, please refer to supplementary.

\subsection{Main Results}

\noindent{\textbf{Human3.6M}}
We evaluate our PaFF model on Human3.6M dataset and compare it with the previous best method~\cite{shin2020multi} and other existing 3D human pose estimation methods, as shown in Table. ~\ref{tab:h36m}. Our PaFF model achieves 33.02 MPJPE improving the previous best method~\cite{shin2020multi} by 29.6\% and 26.9 PA-MPJPE improving by 17.2\%. By comparing the calibration-free PaFF,~\citet{liang2019shape} and~\citet{shin2020multi}, we show PaFF shows the state-of-the-art performance without knowing multi-view cameras.
By comparing with MuVS~\citet{huang2017towards} - a multi-view fitting pipeline using 2D estimated keypoints and 2D silhouette, we show our PaFF can perform better, which implies the advantage of not relying on noisy 2D detection when cameras are few. By comparing with Deep Triangulation~\cite{iskakov2019learnable}, it is seen that PaFF performs close to the direct 3D keypoints estimation.

\noindent{\textbf{MPI-INF-3DHP}}
We evaluate our model on MPI-INF-3DHP to show the generalization of PaFF. As shown in Table ~\ref{tab:mpi}, our model outperforms the other methods in all of the metrics. It does not show a significant improvement as in Human3.6M due to the noisy 3D labels in MPI-INF-3DHP
~\cite{shin2020multi}.

\begin{table}
  \centering
  \begin{tabular}{llll}
    \hline
    Methods     & MPJPE & PCK & AUC \\
    \hline
    \citet{liang2019shape}  & 59.0  & 95.0 & 65.0   \\
    \citet{shin2020multi}     & 50.2 & 97.4 & 65.5   \\
    PaFF(Ours)     & \textbf{48.4} & \textbf{98.6} & \textbf{67.3} \\
    \hline
  \end{tabular}
    \caption{Results on the MPI-INF-3DHP. The higher results of PCK and UAC mean better performances.} \label{tab:mpi}
\end{table}

\subsection{Qualitative Experiments (More in Supplementary)}

\begin{figure}
\centering
\includegraphics[height=0.454\columnwidth,width=1\columnwidth]{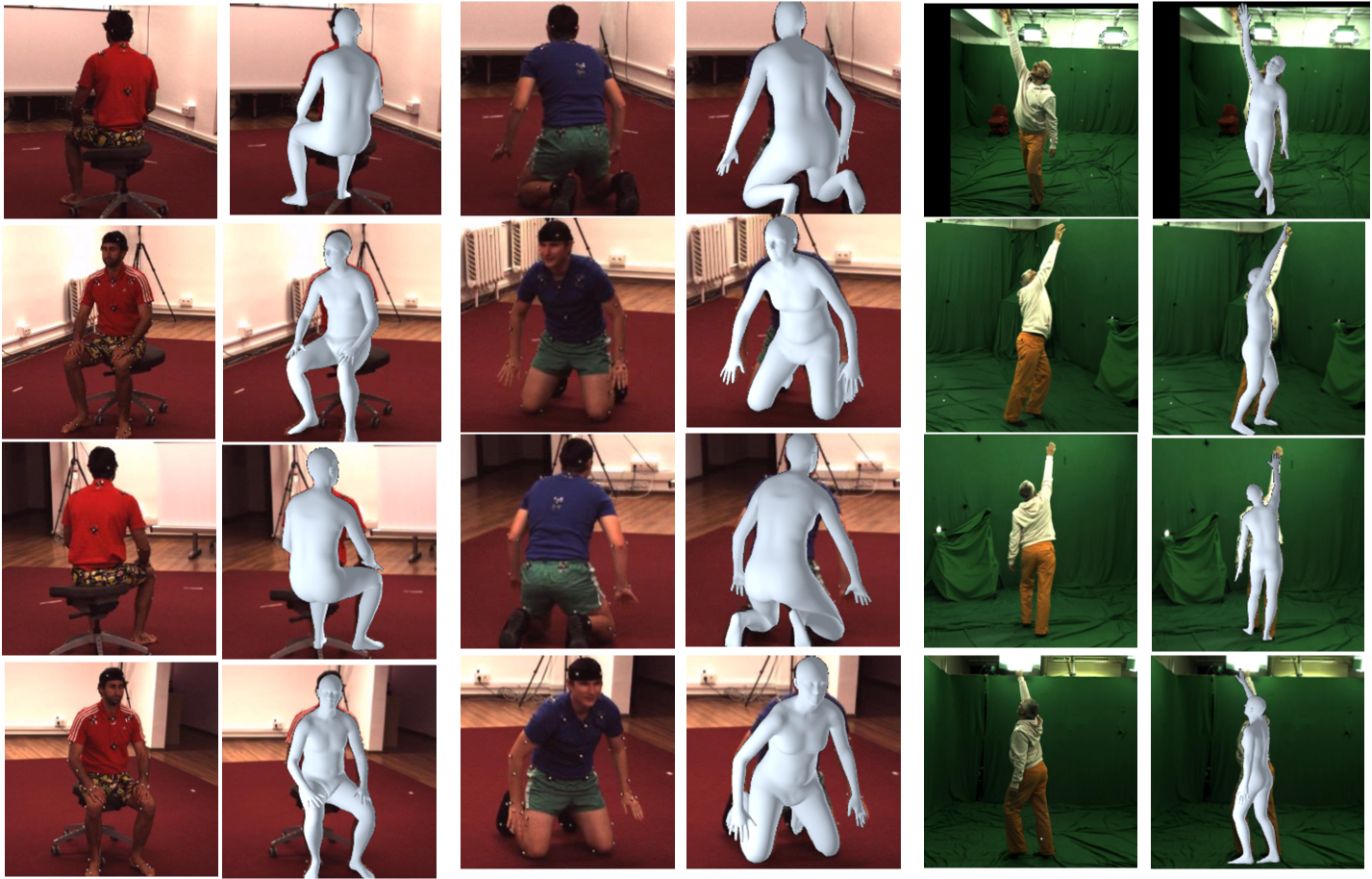}
\caption{Qualitative results on Human3.6M and MPI-INF-3DHP. Each row means a different view.}
\label{fig:vis1}
\end{figure}

\noindent{\textbf{Human Mesh Recovery}} We demonstrate three qualitative examples in Fig.~\ref{fig:vis1} on Human3.6M and MPI-INF-3DHP datasets. Benefiting from the iterative refine process with PaF feedback features, our method could accurately aggregate multi-view information to handle handling object occlusion (chair) and self-contact (the first example). The second and third examples demonstrate the robustness of our PaFF for estimating unusual poses.

\begin{figure}
\centering
\includegraphics[width=0.95\columnwidth]{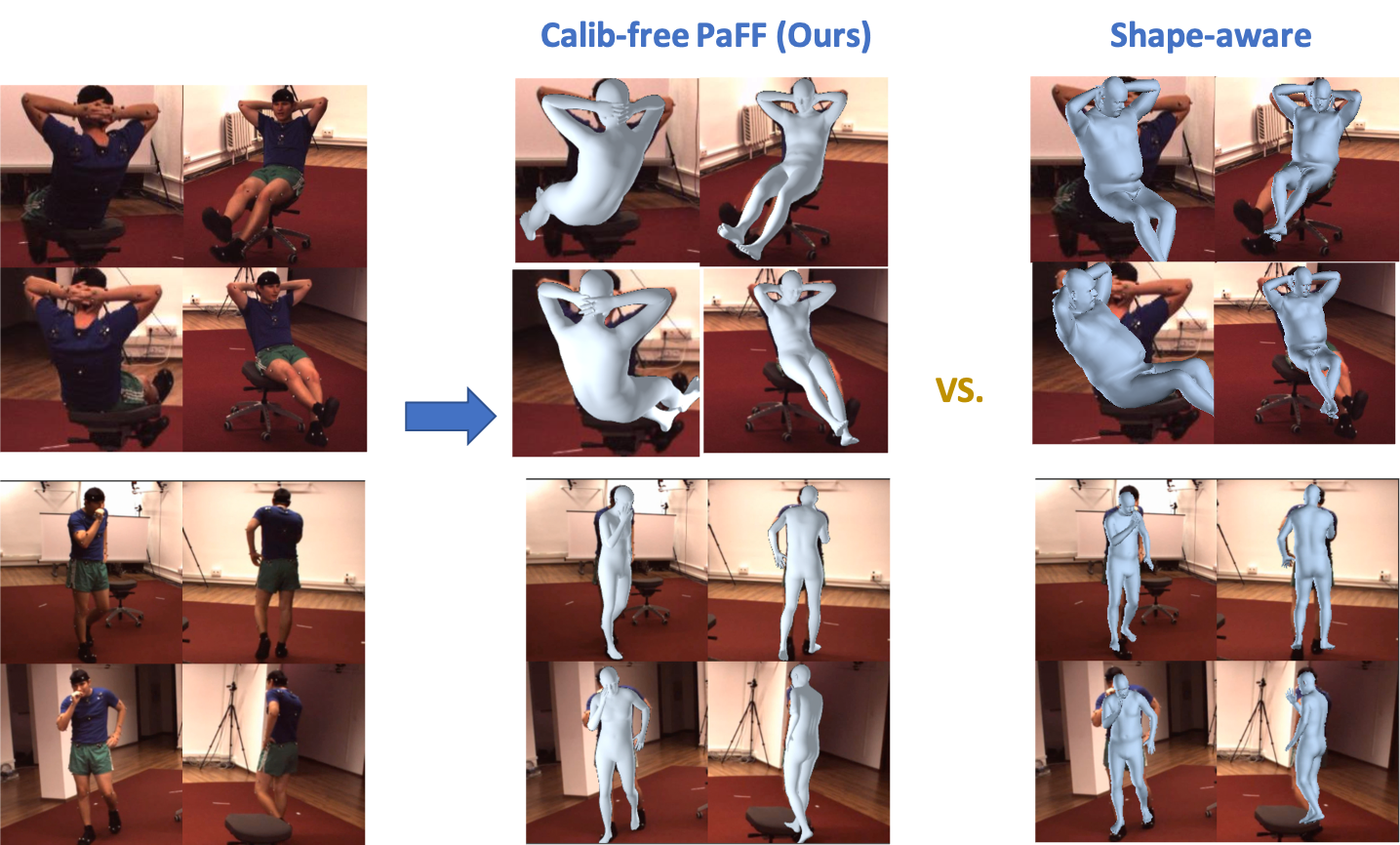}
\caption{Visualization of the effect of Multi-view PaF Feature by comparing the calibration-free PaFF and Shape-aware~\cite{liang2019shape} in two cases.}
\label{fig:vis2}
\end{figure}

\noindent{\textbf{Multi-view PaF Feature Fusion}}
To fairly compare with the calibration-free method~\cite{liang2019shape} (`Shape-aware') in Fig.~\ref{fig:vis2}, we adapt our method to a calibration-free version (`Calib-free PaFF'). Note that our `Calib-free PaFF' could align body better because our multi-view PaF feature fusion can alleviate estimation misalignment.

\subsection{Ablation Study}

\textbf{Multi-view Fusion Architecture Choice for Multi-view Pose \& Shape Estimation}
Fusion using a Transformer encoder for pose \& shape estimation has a lower performance than the fully connected layer + max-pooling in Table. ~\ref{tab:ablation:pose_shape}. The reason for the phenomenon might be that the self-attention mechanism would mess up some views' feedback signals for the multiple pose \& shape parameters.
Moreover, there are additional options for the aggregation function, such as average pooling (AP) and Softmax\_Sum (do softmax first, then sum up). By comparing the performance of these aggregation functions in Table ~\ref{tab:ablation:pose_shape}, we find that the max-pooling operation performs the best, which shows the effectiveness of keeping the maximum essential feedback signals from the feedback features.

\begin{table}
  \centering
  \begin{tabular}{llll}
    \hline
    Methods     & MPJPE & PA-MPJPE & PVE \\
    \hline
    AP & 36.1  & 29.3 & 53.9  \\
    SOFTMAX\_SUM & 34.9 & 28.5 & 51.3   \\
    Transformer + MP & 34.4 & 28.4 & 51.7   \\
    MP(ours)     & \textbf{33.0} & \textbf{26.9} & \textbf{48.9} \\
    \hline
  \end{tabular}
    \caption{Multi-view Aggregation Functions for Pose \& Shape. AP is average pooling. SOFTMAX\_SUM is to do softmax first among multi-views then do summation. MP is max-pooling. Transformer + MP goes through a transformer fusion module first then do max-pooling. } \label{tab:ablation:pose_shape}
\end{table}

\textbf{Different Orientation Estimation Methods}
We compare three different orientation estimation methods as shown in Table.~\ref{tab:ablation:O}.
By comparing `Ind O' and `Ind O + Align', we can see a clear improvement in MPJPE after applying Global O Aligner, which shows the benefit of aligning with camera rotations.
 We find that the performance of the transformer encoder + Global O Aligner structure is superior to the view independent O estimation (`Ind O + Align'), which is due to the ability to renew each view's alignment information with the transformer's cross view attention. The angular O Err is improved from the top to the bottom but with a small step, in which some large orientation angle improvements for some hard cases are averaged in the large data.

\begin{table}
  \centering
  \begin{tabular}{lllll}
    \hline
    Methods     & MPJPE & PA-MPJPE & PVE & O Err \\
    \hline
    Ind O & 44.9  & 28.2 & 65.2 & $6.5^{\circ}$  \\
    Ind O + Align & 34.3 & 27.7 & 49.8 & $5.8^{\circ}$  \\
    Tran + Align (ours) & \textbf{33.0} & \textbf{26.9} & \textbf{48.9} & $\textbf{5.1}^{\circ}$ \\
    \hline
  \end{tabular}
    \caption{Global Orientation Fusion Methods: Ind O is the same with Calib-free PaFF except for using real focal length; Ind O + Align additionally uses Global O Aligner. Tran + Align uses transformer multi-view fusion and Global O Aligner. O Err is in degrees.} \label{tab:ablation:O}
\end{table}

\textbf{Different Translation Estimation Settings}
In Table.~\ref{tab:ablation:T}, by comparing `w./o. Scale' and `w. scale', we have observed clear improvements in PMJPE and PVE after adapting inferred body scale, which proves the advantage of solving scale ambiguity induced by the orthogonal cameras.

\begin{table}
  
  \centering
  \begin{tabular}{lllll}
    \hline
    Methods     & MPJPE & PA-MPJPE & PVE \\
    \hline
    w./o. Scale & 36.5  & 26.9 & 58.6  \\
    w. Scale (ours) & \textbf{33.0} & \textbf{26.9} & \textbf{48.9} \\
    \hline
  \end{tabular}
    \caption{Global Translation Estimation. `w./o. Scale' does not estimate a global translation. `w. Scale' estimates a global translation and an adaptive scale. } \label{tab:ablation:T}
\end{table}


  

\section{CONCLUSIONS}
We present an end-to-end Pixel-aligned Feedback Fusion (PaFF) model to recover a single human mesh from multi-view images. Different from the existing multi-view methods, we extract Pixel Alignment Feedback (PaF) features from images and fuse them with a novel Feedback Fusion Module to infer the misalignment of the current estimation. The feedback fusion module is divided into three fusion modules to disentangle human pose \& shape, global orientation, and global translation in an end-to-end manner. Furthermore, we conduct quantitative experiments on Human3.6M and MPI-INF-3DHP to verify the efficacy of our method, which shows a significant improvement over the previous state-of-the-art. Qualitative results further demonstrate PaFF's potential to deal with the uncommon pose, self-occlusion, and close-contact challenges.
However, the limited multi-view training data can limit the model for instant use in the wild.
In the future, to further improve the performance against the subtle misaligned, a more accurate parametric annotation and a dataset with more diverse human shapes is necessary. Furthermore, our PaFF is an effective and general pipeline, which can be extended to other parametric model regression tasks such as whole-body motion capture~\cite{pymafx2022}, 3D hand reconstruction~\cite{zhou2020monocular,li2022interacting}, and multiple human motion capture~\cite{huang2021dynamic,dong2021fast}.




\section{Acknowledgements}
The work is supported by the National Key R\&D Program of China (Grant No. 2021ZD0113501) and the China Postdoctoral Science Foundation (Grant No. 2022M721844). We would like to thank Yuxiang Zhang, and Mengcheng Li for their help, feedback, and discussions for this paper.

\small{
    \bibliography{main.bib}
}

\end{document}